\documentclass[12pt]{article}
\usepackage{amsmath}
\usepackage{times}
\usepackage{graphicx}
\usepackage{color}
\usepackage{multirow}
\usepackage{url} 

\usepackage[authoryear]{natbib}
\usepackage{soul} 
\usepackage{rotating}
\usepackage{bbm}
\usepackage{latexsym}
\usepackage{amsfonts}
\usepackage{amsmath,bm}
\usepackage{booktabs}

\usepackage{titlesec}
\titleformat*{\section}{\normalsize\bfseries}
\titleformat*{\subsection}{\normalsize\bfseries}

\newcommand{\captionfonts}{\normalsize}

\makeatletter  
\long\def\@makecaption#1#2{%
  \vskip\abovecaptionskip
  \sbox\@tempboxa{{\captionfonts #1: #2}}%
  \ifdim \wd\@tempboxa >\hsize
    {\captionfonts #1: #2\par}
  \else
    \hbox to\hsize{\hfil\box\@tempboxa\hfil}%
  \fi
  \vskip\belowcaptionskip}
\makeatother

\begin{document}

\hspace{13.9cm}1

\ \vspace{20mm}\\

\noindent {\Large Title: Human Eyes Inspired Recurrent Neural Networks are More Robust Against Adversarial Noises} \\

\noindent * This paper has been published in Neural Computation, Volume 36, Issue 9, pages 1713-1743.\\

\noindent \textbf{Minkyu Choi}\\
cminkyu@umich.edu \\
Department of Electrical Engineering and Computer Science, University of Michigan, Ann Arbor, MI 48105, USA

\noindent \textbf{Yizhen Zhang}\\
yizhen.zhang@ucsf.edu \\
Department of Neurological Surgery, University of California, San Francisco, CA 94143, USA

\noindent \textbf{Kuan Han}\\
kuanhan@umich.edu \\
Department of Electrical Engineering and Computer Science, University of Michigan, Ann Arbor, MI 48105, USA

\noindent \textbf{Xiaokai Wang}\\
xiaokaiw@umich.edu \\
Department of Biomedical Engineering, University of Michigan, Ann Arbor, MI 48109, USA

\noindent \textbf{Zhongming Liu}\\
zmliu@umich.edu \\
Department of Biomedical Engineering, University of Michigan, Ann Arbor, MI 48109, USA, and Department of Electrical Engineering and Computer Science, University of Michigan, Ann Arbor, MI 48105, USA\\


\thispagestyle{empty}
\markboth{}{NC instructions}
\ \vspace{-0mm}\\

\begin{center} {\bf Abstract} \end{center}
Humans actively observe the visual surroundings by focusing on salient objects and ignoring trivial details. However, computer vision models based on convolutional neural networks (CNN) often analyze visual input all at once through a single feed-forward pass. In this study, we designed a dual-stream vision model inspired by the human brain. This model features retina-like input layers and includes two streams: one determining the next point of focus (the fixation), while the other interprets the visuals surrounding the fixation. Trained on image recognition, this model examines an image through a sequence of fixations, each time focusing on different parts, thereby progressively building a representation of the image. We evaluated this model against various benchmarks in terms of object recognition, gaze behavior and adversarial robustness. Our findings suggest that the model can attend and gaze in ways similar to humans without being explicitly trained to mimic human attention, and that the model can enhance robustness against adversarial attacks due to its retinal sampling and recurrent processing. In particular, the model can correct its perceptual errors by taking more glances, setting itself apart from all feed-forward-only models. In conclusion, the interactions of retinal sampling, eye movement, and recurrent dynamics are important to human-like visual exploration and inference.


\section{Introduction}

CNNs and brains are different in both front and back ends. In the front end, CNNs use an input layer to evenly sample a grid of pixels. The input size and computational complexity increase quadratically with the dimension of the visual field. In contrast, the brain uses the retina to obtain non-uniform and fisheye-style samples distributed around where the eyes are fixated (\cite{curcio1990human, watson2014formula, bashivan2019neural}). Retinal sampling is much denser and more accurate in the fovea than in the periphery, as the size of the retina's receptive fields increases proportionally with eccentricity, becoming larger as they move away from the fovea (\cite{derrington1984spatial,connolly1984representation,curcio1990human}). This non-linear sampling density allows the brain to keep up with a very wide visual field despite restricted memory and computation (\cite{gattass1981visual,gattass1988visuotopic}). In the back end, CNNs process the input pixels all at once through a single feed-forward pass and arrive at a single perceptual decision per each image input. In contrast, humans may take multiple glances through eye movements and use dual visual streams to guide gaze behavior and support dynamic perception. The magnocellular sub-cortical pathway and its extension onto the dorsal cortical pathway direct where to look at each glance, while the parvocellular sub-cortical pathway and its extension onto the ventral cortical pathway transform the retina samples into abstract representations for recognition (\cite{mishkin1983object, merigan1993parallel}). It is, however, unclear whether and how such front-end and back-end differences explain the gaze behave and perceptual robustness of computer vs. human vision.

Humans use spatial attention to drive eye movements. For rapid recognition, humans tend to look at salient regions while disregarding less informative or insignificant parts. This is because the eyes collect more samples from the fovea such that downstream visual processing is also more dedicated to regions around the fixation. This bias in sampling and processing makes "where to look" an important decision that the brain has to make. To inform this decision, the brain uses a wide view enabled, including both central and peripheral vision, to locate the salient object and use this spatial information as overt attention to direct eye movement (\cite{deubel1996saccade, wiecek2012effects}). This process involves the brain's "where" pathway and the oculomotor control pathway (\cite{colby1999space, rizzolatti2003two, corbetta2002control}). 

Under normal viewing conditions, humans move their eyes to collect more information and refine visual perception. Human vision is robust, unlike computer vision. A small amount of adversarial noises can deceive computer vision, but appears trivial to humans (\cite{goodfellow2014explaining}). However, when humans have an extremely restricted time (e.g., 70ms or less) to observe an image, the brain may also make perceptual mistakes given the same adversarial noises (\cite{elsayed2018adversarial}). In this case, humans can only afford a single glance while the brain is limited to its feed-forward processing, rather than feedback or recurrent processing (\cite{lamme2000distinct}). The brain's feedforward visual processing takes ~150 ms to reach a perceptual decision (\cite{thorpe1996speed}) for rapid object recognition (\cite{dicarlo2012does}), akin to the mode of operation in CNNs. Therefore, it is likely that eye movement, feedback and recurrent processing are necessary for robust human vision. The retina may place a key role in adversarial robustness. Depending on the distance from the point of fixation, or eccentricity, the same object bears different retinal patterns, when it is in the fovea vs. periphery (\cite{bouma1970interaction, lettvin1976seeing, rosenholtz2016capabilities, balas2009summary, stewart2020review}). This dynamic sampling acts as a form of data augmentation, possibly making visual perception less vulnerable to minor distortions or adversarial perturbations (\cite{vuyyuru2020biologically}).

Therefore, the retina has profound impacts on human vision, but rarely explored in computer vision. In this study, we draw inspirations from human eyes, dorsal and ventral streams, and spatial attention to design a recurrent neural network for computer vision, and evaluate the model's attention and adversarial robustness. The model includes three modules: an input sampler that takes retinal samples around the fixation, an attention network that mimics the brain's dorsal visual pathway and guides where to look next, and a recognition network that mimics the ventral visual pathway and represents the retinal samples recurrently for object recognition. As illustrated in Fig.~\ref{fig:models}, we design three variations of such a model. Each of them uses a different strategy for sampling visual input with respect to the point of fixation, including cropping image patches with a single field of view, with two fields of view, or applying retinal foveation and non-uniform sampling. These models all attempt to mimic the brain's ability to engage in saccadic eye movement and recurrent neural processing (\cite{lamme2000distinct, kar2019evidence}). We hypothesize that the model using eye-like, retinal transformation learns attention behavior that is closer to human attention. Additionally the model with retinal transformation can progressively refine the perceptual decision and improve robustness against adversarial perturbations as it allocates more time to take additional glances at an image under attack, allowing for a more refined perceptual decision.

\section{Related Works}

\subsection{Two-Stream Architecture}
In the field of computer vision, focusing on distinct aspects of data, such as spatial and temporal features, often proves beneficial for solving specific tasks like object recognition and action recognition. To handle these distinct features, two or more parallel processing streams specialized for each feature can be employed.

In object recognition, \cite{esteves2017polar, sermanet2014attention, wang2020glance, guo2019global} utilize dual-stream architectures to support both global and local feature processing. In these works, one stream processes the entire image area, extracting global features that provide a broad understanding of the scene. Based on these global features, the models determine where to allocate more resources to obtain detailed local features. Subsequently, the other stream processes these local features, allowing for more precise object recognition. By processing only the informative image areas, these models save computational costs while achieving comparable recognition accuracies to single-stream models with larger sizes. For video processing, dual-stream architectures specialized for temporal and spatial information processing demonstrate improved action recognition performance (\cite{simonyan2014two, feichtenhofer2017spatiotemporal, wang2016temporal}).
In work by \cite{choi2023dual}, the authors proposed two-stream architectures to model the dorsal and the ventral visual streams of human brains. The two streams of the model are trained to perform distinct functions of saliency prediction and object classification. The results demonstrate that the two streams in the model resemble the representations of the brain's dorsal and ventral visual streams, respectively. 

Our study also employs a two-stream architecture for local and global feature processing. 
However, unlike previous works that simply crop patches from the given images (\cite{sermanet2014attention, wang2020glance, guo2019global}), our model uses retina-like image sampling, enabling the model to effectively capture detailed local features while not ignoring the global context of the images. 
Our work shares similarities with the study by \cite{choi2023dual} in terms of utilizing two streams and retinal transformation. However, a key distinction lies in our approach to modeling visual attention. Unlike \cite{choi2023dual}, which employs a human saliency dataset to mimic human attention, our model is exclusively trained for object recognition without relying on any human saliency dataset. Consequently, in our model, the emergence of human-like visual attention occurs as a natural by-product of the object recognition task, rather than being directly trained to replicate human attention patterns.

\subsection{Foveated Visions} 
In primate vision, the visual acuity is highest at the center of the gaze, a region called the fovea, due to the dense concentration of cone photoreceptors. This phenomenon, known as foveated vision, enables high-resolution and color perception in the central visual field. As the distance from the fovea increases, visual acuity decreases, reflecting the natural distribution of photoreceptors in the retina (\cite{curcio1990topography, curcio1990human, weber2009implementations}). 
The concept of foveated vision has been extensively explored in the field of machine vision (\cite{deza2020emergent, cheung2016emergence, pramod2022human}). Researchers have investigated its applications across various domains, such as image rendering for virtual reality, predicting human scanpaths, and improving object recognition performance.

Specifically, in virtual reality, low-resolution rendering of peripheral image regions based on foveated vision has been shown to significantly reduce rendering latency due to the reduced accuracy of human peripheral vision. Several studies have explored this approach and demonstrated its effectiveness in reducing computational requirements while maintaining the quality of the rendered image (\cite{jabbireddy2022foveated, kaplanyan2019deepfovea}).

Furthermore, the foveated vision has been widely used in predicting human scanpaths by analyzing eye movement patterns during visual perception. Researchers have developed models that can predict the order in which humans fixate on different regions of an image or scene, which can be useful in various applications such as advertising, design, and robotics (\cite{wang2017scanpath, berga2020modeling, bao2020human}).

In addition to predicting scanpaths, foveated vision has been shown to improve object recognition performance or reduce the computational costs of recognition systems. By using a high-resolution foveal region to focus processing resources on important areas of the image and a low-resolution peripheral region for less important areas, recognition systems can achieve higher accuracy with lower computational requirements. This approach has been explored in several studies and has shown promising results in improving recognition accuracy while reducing computational costs (\cite{thavamani2021fovea, min2022peripheral, jonnalagadda2021foveater, wang2021use}).

\subsection{Recurrent Attention}
A crucial difference between human and machine vision is that humans explore images or scenes by directing attention with eye movements (overt attention) or without eye movements (covert attention). Prior works have attempted to make machines solve downstream tasks by teaching them where to look through a recurrent process. 
Given an attention focus (or the fixation), the attended region can be cropped at various resolutions or scales (\cite{wang2020glance,xu2015show, mnih2014recurrent, sermanet2014attention}). Arguably better than hard cropping, the attended region may be subject to retinal transformation (\cite{vuyyuru2020biologically, bashivan2019neural, akbas2017object}) or polar transformation (\cite{esteves2017polar}) inspired by the primate retina (\cite{curcio1990human, thibos1987retinal, geisler1986sampling, coletta1987psychophysical}). For covert attention, models generate and apply soft weightings for all features and locations depending on their relative importance (\cite{xu2015show, zoran2020towards,jaegle2021perceiver}). 

These recurrent attention models are similar to the human brains (\cite{kietzmann2019recurrence}). Because of this similarity, understanding recurrent attention models would provide insights into the mechanisms of adversarial robustness in human vision. However, despite their importance, recurrent attention models have rarely been the focus of research compared to their counterparts, feedforward CNNs. In the current work, we design and test different recurrent attention models, and show how they are affected by adversarial noise as more recurrent steps are deployed.

\subsection{Adversarial Attacks}
An adversarial attack refers to an attempt to deceive machine learning models by adding carefully designed perturbations to an input image (\cite{goodfellow2014explaining, szegedy2013intriguing, carlini2017towards, madry2017towards, athalye2018synthesizing}). The perturbation, known as adversarial noise, can be optimized with the projected gradient descent (PGD) (\cite{madry2017towards}), fast gradient sign method (\cite{goodfellow2014explaining}) and  Carlini \& Wagner attack (\cite{carlini2017towards}), among others. To counteract adversarial noise, defensive methods have been proposed (\cite{goodfellow2014explaining, gu2014towards, papernot2016distillation, xie2017mitigating}) but still remain vulnerable (\cite{athalye2018synthesizing, carlini2017towards, he2017adversarial}). 

Unlike computer vision models, humans do not seem to suffer from the same vulnerability, especially when enough time is available for humans to observe the image under attack (\cite{elsayed2018adversarial}). This distinction has motivated prior works to take inspiration from human vision for computer vision. For example, \cite{luo2015foveation} have found that allowing models to fixate on different image regions can alleviate the effect of adversarial noise. 
More recently, \cite{dapello2020simulating} shows that attaching a block with properties of V1 at the front of feed-forward CNNs, can make CNNs more robust. \cite{huang2020neural} demonstrate that a neural network with predictive coding improves adversarial robustness. In the study of \cite{berrios2022joint}, authors show that their adversarially robust model achieves higher explainable variance for some brain areas compared to the biologically plausible neural networks, showing the possible link between the adversarially trained transformer and the explainability in the brains. 

\cite{vuyyuru2020biologically} demonstrate that non-uniform spatial sampling and varying receptive fields that mimic the retinal transformation in the primate retina can also improve the robustness against adversarial attacks. \cite{harrington2021finding} show that the representation robust against adversarial noises is more attributable to processing information from the periphery, as opposed to the fovea. 

In line with the related work, we also explore biologically inspired computational mechanisms for adversarial robustness. Similar to \cite{vuyyuru2020biologically}, we use a retina-like front-end to generate fixation-dependent retinal input for image recognition. A notable distinction is that \cite{vuyyuru2020biologically} prefixes the points of fixation, whereas, in our study, the fixation is adaptive and sequentially inferred via an attention module that learns where to look next. As such, our model iteratively and sequentially samples an image into time-varying retinal patterns, similar to retinal input to the brain during saccadic eye movement.

\begin{figure}[h!]
  \centering
  \includegraphics[width=0.99\linewidth]{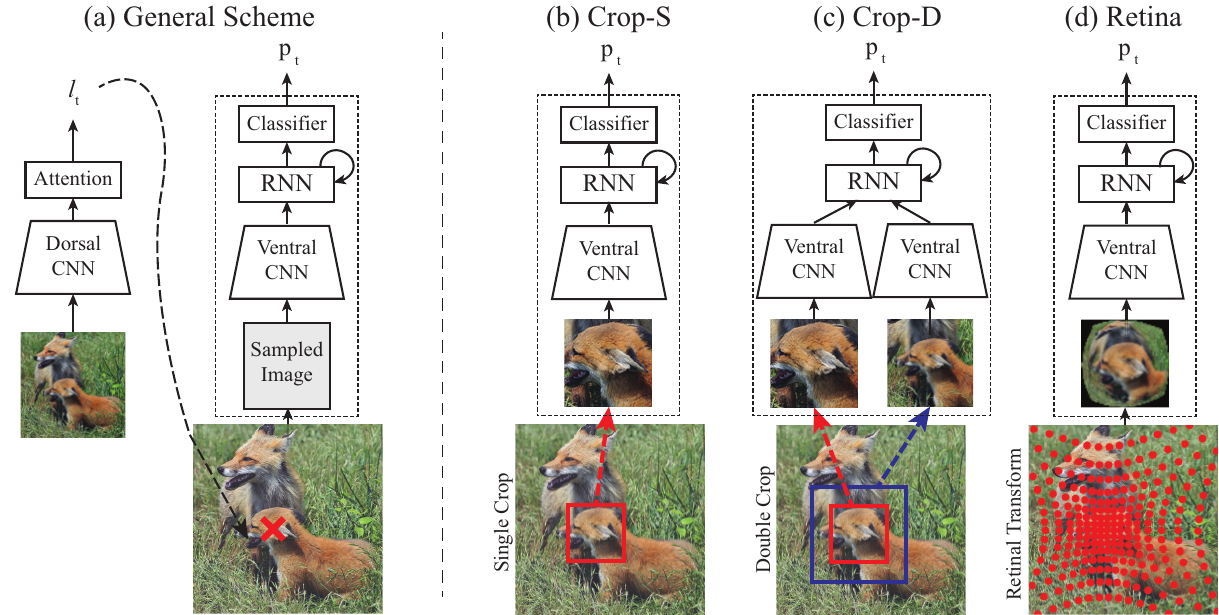}
  \caption{
Recurrent models for eye movement and visual recognition. (a) The general two-stream model architecture: The model includes two streams - the dorsal (left) and ventral (right) streams. The dorsal stream has a broad field of view and generates a fixation point ($l_{t}$) at a step (or glance). The ventral stream takes selective samples around the fixation, extracts a representation, and accumulates the representations across multiple glances for object recognition ($p_t$). Using this general scheme, we design and test three different implementations of the ventral stream, namely Crop-S, Crop-D, and Retina, illustrated in (b) through (d). (b) Crop-S crops a small image region (shown as the red box) around the fixation. (c) Crop-D crops two regions (shown as the red and blue boxes) around the fixation and samples them with different resolutions such that the same number of samples is extracted from either region. (d) Retina applies retinal transformation and extracts non-uniform samples with respect to the fixation. Those models use the same architecture for the dorsal stream, for which the weights are learned separately alongside the different ventral stream models. 
  }
  \label{fig:models}
\end{figure}

\section{Method}
In this section, we detail the methodology behind our approach, which integrates elements of human visual processing into our computer vision models. 
As illustrated in Fig.~\ref{fig:models}(a), the dorsal stream has a wide field of view that covers the whole image sampled with lower resolution. It learns spatial attention, predicts where to look next, and passes the predicted fixation ($l_{t}$) to the ventral stream. The ventral stream has a narrow field of view around the fixation, samples the input, learns to represent the samples for each fixation,  recurrently accumulates the representation across different fixations, and outputs the probability in image classification ($p_t$). 

Specifically, the dorsal stream includes two modules: \texttt{DorsalCNN} and \texttt{Attention}, and the ventral stream includes \texttt{VentralCNN}, \texttt{RNN}, and \texttt{Classifier}. Three model variations shown in Fig.~\ref{fig:models} (b), (c), (d) all share the same architecture for the dorsal stream, but use different sampling strategies for the ventral stream: a single-cropped field of view (Crop-S), double-cropped field of views with different resolution but the same matrix size (Crop-D), and retinal sampling (Retina). Further details of these configurations are provided below..

\subsection{Image Sampling} \label{sec:image_sampling}

\noindent \textbf{Single Crop: } The model Crop-S crops a rectangular patch around the fixated location as the input to the ventral stream (Fig~\ref{fig:models} (b)). 

\noindent \textbf{Double Crop: } The model Crop-D processes two rectangular patches around the same fixation point, as shown in Fig~\ref{fig:models} (c), with different scales and resolution. They are further resized to have the same patch size.

\noindent \textbf{Retinal Sampling: } 
We use retinal sampling with the knowledge of biological visual systems. In primate visual systems, both retinal ganglion cells and neurons at early visual areas have increasingly larger receptive fields yet lower resolution at higher eccentricity relative to where the eyes are fixated in the visual field (\cite{gattass1981visual,gattass1988visuotopic, freeman2011metamers}). In addition, more cells and neurons are devoted to the central vision than to peripheral vision (\cite{curcio1990human, watson2014formula, bashivan2019neural}). Inspired by these properties, we design an input layer with two steps: 1) foveated imaging and 2) non-uniform sampling. 

Foveated imaging (\cite{duchowski2004gaze,perry2002gaze}) varies image resolution and acuity along eccentricity: higher resolution around the fixation and progressively lower resolution towards the periphery. We implement this by applying a larger Gaussian kernel to peripheral regions but a smaller kernel to the central region, resulting in greater smoothness in the periphery. More details are included in Appendix A. Since each foveated image is transformed from the original image with a varying extent of spatial blurring, the feed-forward convolutional layers  effectively have eccentricity-dependent receptive fields. This effect is similar to how biases from different receptive fields in the retina are passed to downstream visual areas. 

Retinal sampling collects non-uniform discrete samples from the foveated image. Fig.~\ref{fig:models}(d) shows an example of retinal sampling points and the resulting retinal image. 
Suppose that we want to sample an \(N\times N\) foveated image to fill an \(n\times n\) retinal image with \(n<N\). We first calculate the eccentricity $e (x,y;f_x,f_y)$ of any pixel location $(x,y)$ with respect to the fixation point $(f_x,f_y)$ in the foveated image (Eq.~\ref{eq:eccentricity1}). Similarly we calculate $\rho (i,j;f_i,f_j)$ as the distance from fixation $(f_i,f_j)$ to pixel $(i,j)$ (Eq.~\ref{eq:eccentricity2}) in the retinal image.
\begin{align}
    e (x,y;f_x,f_y) &= \| (x,y) - (f_x,f_y) \|_2     \label{eq:eccentricity1}\\ 
    \quad \rho (i,j;f_i,f_j) &= \| (i,j) - (f_i,f_j) \|_2     \label{eq:eccentricity2}
\end{align}

We then relate the eccentricity in the foveated image \(e\) to \(\rho\) of the retinal image through a non-linear mapping function \(g(\rho)\) (Eq.~\ref{eq:g_rho}).

\begin{equation}
    e = g(\rho) =   \frac{N\sinh(\rho \cdot b \frac{2}{N})}{2\sinh (b \frac{n}{N})}  
    \label{eq:g_rho}
\end{equation}

\begin{figure}
  \centering
  \includegraphics[width=0.9\linewidth]{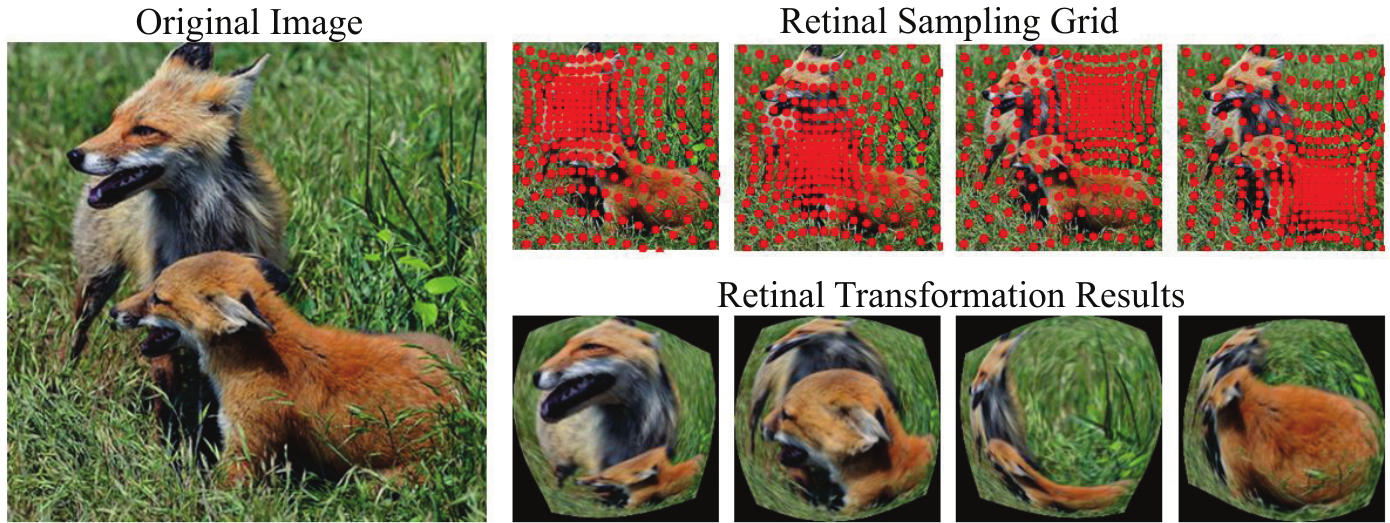}
  \caption{
  Examples of the retinal transformation. Left: Original image, Right: Retinal sampling grid and resulting retinal images.
}
  \label{figs:bs}
\end{figure}

Here, \(b \in \mathbb{R}^+\) is a hyper-parameter. Its value controls the degree of non-uniform sampling, which is set to $12.0$ in our work.
When there is no retinal sampling or $b$ is very small, it can be considered that an image is under the fovea. Therefore, visual acuity is high everywhere sampled. As $b$ becomes larger, the fovea area decreases, and the periphery increases, while the retinal image is increasingly distorted relative to the original image (See Fig.~\ref{figs:bs} and Fig.~\ref{fig:vis_attn}(c) for examples).

\begin{figure}
    \centering
    \includegraphics[width=0.6\linewidth]{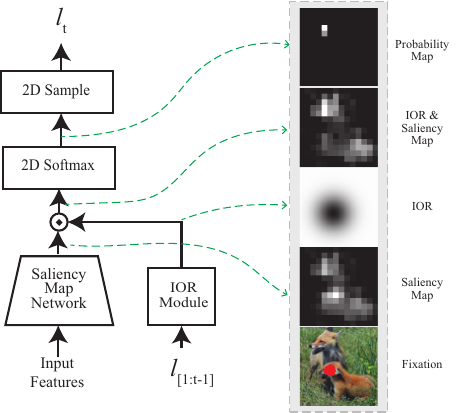}
    \caption{Details of the \texttt{Attention} module in the dorsal stream. \texttt{Attention} produces 2D saliency map and fixation point at time step $t$ ($l_{t}$). Intermediate representations from each module are shown on the right. }
    \label{fig:attn}
\end{figure}

\subsection{Dorsal Stream}
In the preceding section, we explained how we implement retinal sampling to the visual input when a fixation point is present. Now, the next question is: how do we integrate an overt attention mechanism to determine where to look? Inspired by human visions, this process is facilitated by the dorsal stream in our proposed model, which is designed to generate a sequence of fixation points on the given images.

The dorsal stream includes \texttt{DorsalCNN} and \texttt{Attention} module. The input to the dorsal stream is an image with the full view, but in a low resolution. The \texttt{DorsalCNN} is a stack of convolutional layers. 
\texttt{Attention} determines the next location to focus on. It uses the feature maps from all layers from \texttt{DorsalCNN}, after they are resized to the same size and then concatenated along the channel dimension. 
Fig.~\ref{fig:attn} shows the architecture of the attention module. To predict the next fixation point (\(l_{t}\)), the module maps salient regions with a \texttt{Saliency Map Network}. It includes a convolutional layer with \(3\times 3\) kernels and outputs a 2D saliency map (\cite{itti1998model, koch1987shifts}), highlighting the candidate regions for the next fixation. 

To prevent future fixations returning to the previously attended regions, the inhibition-of-return (IOR) module ($\texttt{IOR-Module}$) (\cite{itti2001computational,posner1984components,akbas2017object,najemnik2005optimal}) reduces the saliency of previously attended regions. Specifically, the IOR at time $t$ is expressed as below.  

\begin{equation}
    \textbf{ IOR}(t) = \texttt{IOR-Module}(\bm{l}_{1:t})  = \textbf{ReLU} \Big( \bm{1} - \sum_{\tau=1}^{t} G(\bm{\mu}=\bm{l}_{\tau}, \bm{\Sigma} =\sigma^2\bm{I}) \Big)
    \label{eq:IOR}
\end{equation}

Here, $G$ is a normalized 2D Gaussian function centered at \(\bm{l}_{\tau}\) with a standard deviation \(\sigma\) at the \(\tau\)-th step. Its values are normalized so that the maximum equals 1. We sum the normalized Gaussian kernels across all previous time steps to prohibit future fixations from going back to the previously visited locations. Subtracting the accumulated Gaussian kernels from a matrix of ones creates a soft mask with high values at unattended regions but low values at previously attended regions. Applying this mask to the saliency map by element-wise multiplication followed by softmax gives rise to a 2D probability map based on which the next fixation is randomly sampled. Fig.~\ref{fig:attn} shows typical examples of the saliency map, IOR, and the prediction of the next fixation point.

The dorsal stream is trained with reinforcement learning based on the \texttt{REINFORCE} algorithm (\cite{williams1992simple}). At time $t$, the fixation \(\bm{l}_{t}\) generated by the attention module results in a new class prediction \(p_{t}\) by the recognition pathway (ventral stream). The reward \(r_{t}\) of choosing \(\bm{l}_{t}\) as the fixation is calculated as the reduced classification loss relative to the previous time step \(r_{t} = CE(p_{t-1}, \text{label}) - CE(p_{t}, \text{label})\), where \(CE\) is the cross-entropy loss. The goal of reinforcement learning is to maximize the discounted sum of rewards, \(R = \sum_{t=1}^T \gamma^{t-1} r_t \), where \(\gamma \in (0, 1)  \) is the discount factor and it is set as 0.8.

\subsection{Ventral Stream}
The recognition pathway includes \texttt{VentralCNN}, \texttt{RNN} and \texttt{Classifier} stacked as shown in Fig.~\ref{fig:models}. \texttt{VentralCNN} consists of a stack of convolutional layers to extract features from the retinal samples at each time step. 
The extracted features initially go through global average pooling, and then are channeled to a recurrent neural network (RNN) equipped with gated recurrent units (\cite{chung2014empirical}). The \texttt{RNN} learns to accumulate information across different fixation points, which are then used for object recognition at the \texttt{Classifier} (a fully connected layer) at every time step $t$. The learning objective is to minimize the cross entropy losses summed across \(T\) time steps.

\subsection{Implementation Details}
Convolutional layers in \texttt{VentralCNN} and \texttt{DorsalCNN} have the same architecture (but distinct parameters), and they all have convolutional layers using \(3\times 3\) kernels with stride equal to $1$.  \texttt{VentralCNN} and \texttt{DorsalCNN} have \(\texttt{Conv}(64) \times 2 \to \texttt{MaxPool} \to \texttt{Conv}(128)\times 2  \to \texttt{MaxPool} \to \texttt{Conv}(256)\times 2 \to  \texttt{MaxPool}  \to \texttt{Conv}(512)\times 2\), where \(\texttt{Conv}(C) \times k\) represents $k$ convolutional layers with $C$ channels. \texttt{MaxPool} has a kernel size of $2$ and a stride of $2$. The sampled images for \texttt{VentralCNN} and \texttt{DorsalCNN} are in the size of \(64 \times 64\) and \(128 \times 128\) respectively. The dorsal and the ventral stream as shown in Fig.~\ref{fig:models} are trained together.  For the Crop-D model, the same \texttt{VentralCNN} sharing the weights is used to extract features from the both patches, and the resulting features are summed to be forwarded to the next module. Code and data are publicly available \footnote{\url{https://github.com/minkyu-choi04/rs-rnn}}.

\subsection{Training Details}
We train these three two-stream models (Crop-S, Crop-D and Retina) in three stages. First, we train models on single-label classification on ImageNet100. 
We randomly sample $100$ classes from ILSVRC2012 (\cite{deng2009imagenet}) with $1,000$ classes in total to form ImageNet100 to save computational costs. Then, the models are trained on multi-label classification tasks on MS-COCO (\cite{lin2014microsoft}). We include MS-COCO in the training process because of the complexity of the dataset. Unlike ImageNet, images in MS-COCO include multiple objects, and the sizes of the objects are usually smaller than those of ImageNet, which is beneficial for training the dorsal stream for mimicking human attention. In the last stage, we fine-tune the models on ImageNet100 again. This fine-tuning stage is required to test our models on adversarial attacks because most of the adversarial attack algorithms are for single-label images. 

In the first stage on ImageNet100, we train our models for $90$ epochs using Adam optimizer (\cite{kingma2014adam}) (lr=$0.002$, $\beta_1$=0.9, $\beta_2$=0.99). The learning rate is decreased by $10$ at $30$th, $60$th and $80$th epochs. During training, models are allowed to deploy four fixations to explore images. 
In the second stage on MS-COCO, we train our models for for $90$ epochs using Adam optimizer (lr=$0.001$, $\beta_1$=0.9, $\beta_2$=0.99). The learning rate is decreased by $10$ at $30$th, $60$th and $80$th epochs. During training, models are allowed to deploy eight fixations to explore images. 
In the last stage on ImageNet100, we train our models for for $20$ epochs using Adam optimizer (lr=$0.0005$, $\beta_1$=0.9, $\beta_2$=0.99). The learning rate is decreased by $10$ at $10$th epoch. During training, models are allowed to deploy four fixations to explore images.

\section{Experiments}
With the proposed models, we first test the performance of the models on single-label and multi-label classification tasks. Then, we test how close the model attention is to human attention. At the same time, the effect of input sampling (retinal transformation, double crops, or single crop) on the generated attention is investigated. 
Lastly, we then test our models on the adversarial attacks to evaluate their adversarial robustness. 
As additional baseline models, we also trained a model with feed-forward CNN (FF-CNN) and a model with recurrent attention, but without overt eye movements (S3TA (\cite{zoran2020towards})). In the literature by \cite{zoran2020towards}, S3TA is trained with adversarial training. However, since our focus is on the model architecture, instead of learning strategies, we use S3TA trained on single-label classification and multi-label classification tasks without adversarial training.

\begin{table}[]
  \caption{Summary of the models.}
  \label{table:sum}
  \centering
  \begin{tabular}{@{}c@{\hskip 0.5in}c@{\hskip 0.5in}c@{\hskip 0.5in}c@{\hskip 0.5in}c@{}}
    \toprule
    Model       & \begin{tabular}[c]{@{}c@{}}ImageNet100 \\ Top-1 Accs\end{tabular} & \begin{tabular}[c]{@{}c@{}}MS-COCO \\ F1 Scores\end{tabular} & Input Type     & Attention \\ 
    \midrule
    Retina      & 76.6\%    & 59.5          & Retinal Image  & Overt     \\
    Crop-D      & 81.2\%    & 57.0          & Double Crop    & Overt     \\
    Crop-S      & 75.5\%    & 47.6          & Single Crop    & Overt     \\
    S3TA        & 82.3\%    & 58.8          & Whole Image    & Covert    \\
    FF-CNN      & 80.8\%    & 57.0          & Whole Image    & -         \\ 
    \bottomrule
  \end{tabular}
\end{table}

\subsection{Model Performances on Classification Tasks}
With the models trained on ImageNet100 and MS-COCO, we report their top-1 accuracy and F1 scores on the validation sets, which were not used for training the models. Table~\ref{table:sum} includes the top-1 accuracy on ImageNet100’s validation set and F1 scores for multi-label recognition on MS-COCO's validation set for all models.  Table~\ref{table:sum} also summarizes the attention properties of all models compared. For the ImageNet100 dataset, models that receive full images (FF-CNN and S3TA) are generally better than the models with sampling images (Retina, Crop-D, and Crop-S), which may be attributable to the large size of the objects in the image. 
For image-sampling models (Retina, Crop-D, and Crop-S), a single glimpse may not capture the entire object. In contrast, FF-CNN models view the entire object, which contributes to their better performance.
However, for MS-COCO, the model Retina's F1 score surpasses the other models, which implies that the complex image datasets with multiple small objects, such as MS-COCO, are more suitable for the models with human-like image sampling strategy, compared to simpler datasets with a single large object, such as ImageNet.

\begin{figure}
  \centering
  \includegraphics[width=1.0\linewidth]{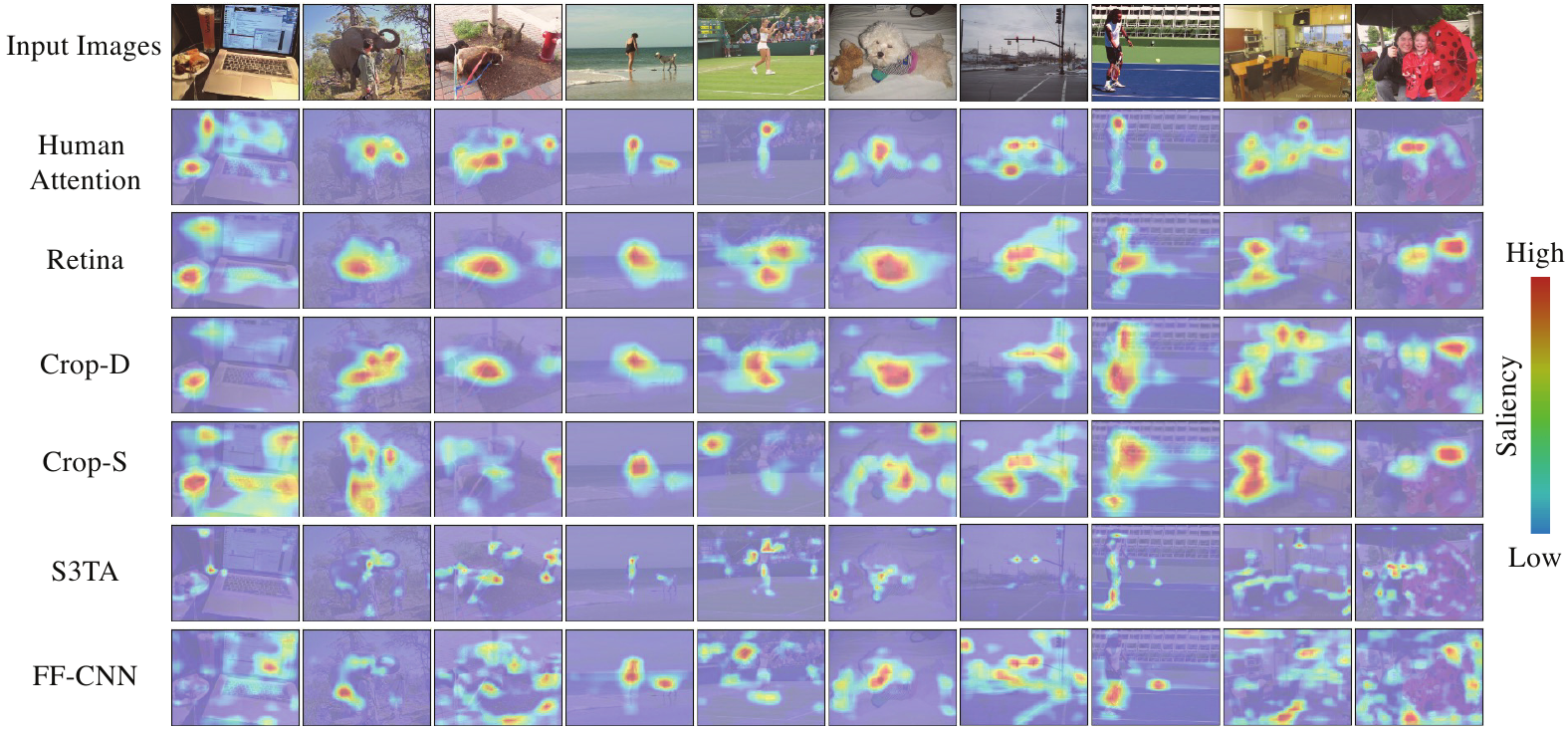}
  \caption{
  Saliency maps from top to bottom: humans, Retina, Crop-D, Crop-S, S3TA, and FF-CNN. Regions of higher saliency are highlighted in red, while areas of lower saliency are depicted in blue.
  }
  \label{fig:attention_maps}
\end{figure}

\subsection{Model Attentions Compared to Human Attentions}
In this batch of experiments, we quantitatively and qualitatively compare the generated attention maps with human attention. 
The proposed two-stream models, particularly the Retina model, successfully capture human-like attention without direct supervision from human saliency data.
The models with explicit attention mechanisms (Retina, Crop-S, Crop-D, and S3TA) are trained to produce and utilize the attention maps for object classification tasks.
Although FF-CNN lacks explicit attention mechanisms, we use Class Activation Map (CAM, \cite{zhou2016learning}) to highlight image regions contributing to object label predictions.
Fig.~\ref{fig:attention_maps} illustrates the attention maps from humans, Retina model, Crop-D, Crop-S, S3TA, and FF-CNN from the top row using the images from the validation set of SALICON (\cite{jiang2015salicon}). In the visualized attention maps, the Retina model and Crop-D largely overlap with the human attention maps. Attentions from S3TA are more selective and fragmented, but still overlap with human attention. However, the attended areas from FF-CNN are fragmented and sometimes focus on non-essential parts of the images.

\begin{table}[]
  \caption{Saliency prediction metrics on SALICON dataset.}
  \label{table:salicon}
  \centering
\begin{tabular}{ccccc}
\hline
       & AUC $\uparrow$            & NSS $\uparrow$            & SIM $\uparrow$            & CC $\uparrow$\\ \hline
Retina & \bm{$0.77\pm<0.01$} & \bm{$0.41\pm<0.01$} & \bm{$0.49\pm<0.01$} & \bm{$0.44\pm<0.01$}  \\
Crop-D & $0.75\pm<0.01$ & $0.38\pm0.01$ & $0.48\pm0.01$ & $0.41\pm0.01$\\
S3TA   & $0.72\pm<0.01$ & $0.29\pm0.01$ & $0.46\pm<0.01$ & $0.29\pm<0.01$  \\
Crop-S & $0.65\pm0.02$ & $0.23\pm0.04$ & $0.40\pm0.02$ & $0.26\pm0.03$  \\
FFCNN  & $0.62\pm<0.01$ & $0.21\pm0.01$ & $0.40\pm<0.01$  & $0.20\pm0.01$  \\ \hline
\end{tabular}
\end{table}

Quantitative evaluation of attention accuracy aligns with the visual observation. We report quantitative metrics of AUC (Area Under the Curve), NSS (Normalized Scanpath Saliency), SIM (Similarity Metric), CC (Linear Correlation Coefficient) in Table~\ref{table:salicon} to measure how close the produced attentions from the models are to the human attentions. The result shows that the attention from the Retina model is closest to the human attention, and Crop-D is the second closest compared to the other models. This implies that the input sampling schemes, with varying resolutions and resembling fovea-periphery, might be helpful in producing attention maps that are similar to human attention. On the other hand, FF-CNN produces the least human-like attention maps.

\begin{figure}
  \centering
  \includegraphics[width=1.0\linewidth]{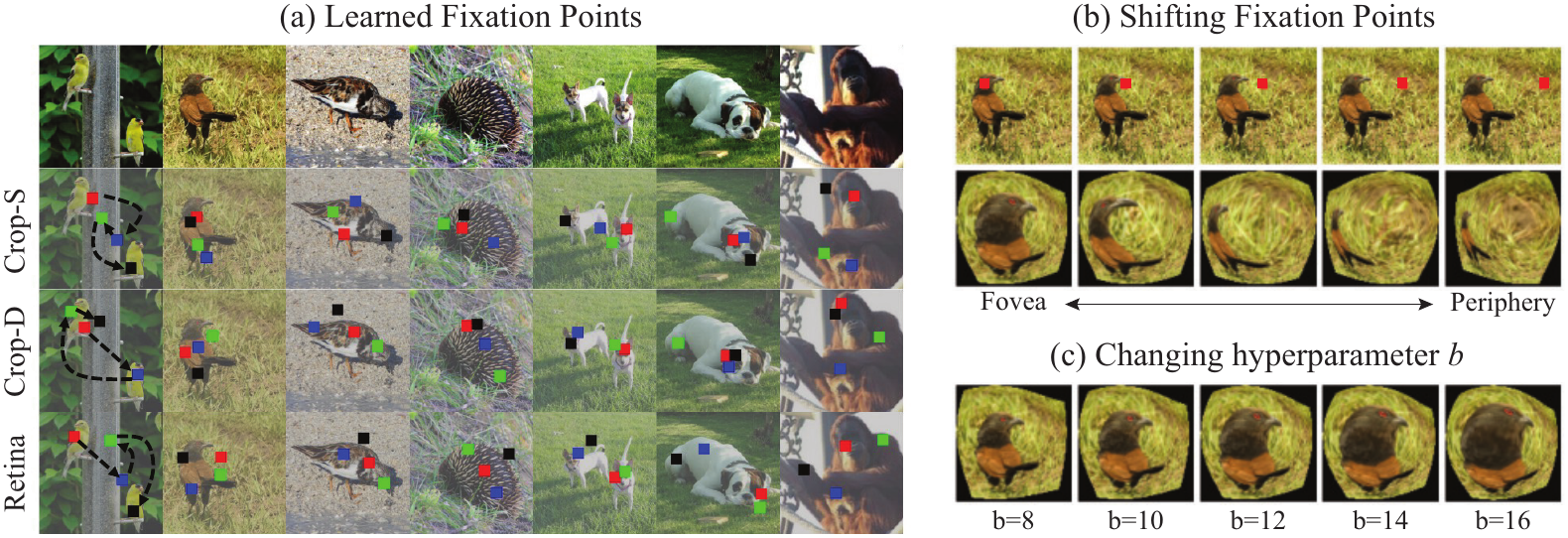}
  \caption{
  (a) Learned fixations from the first four steps. From the first fixation to the last fixations are marked as Red-Blue-Green-Black. First column presents exemplar attention shifts. (b) Visualization of the retinal transformation as the object is shifted from the fovea to the periphery. (c) Examples of the retinal transformation as the hyperparameter $b$ is changing from $8$ to $16$. 
  }
  \label{fig:vis_attn}
\end{figure}

To validate the plausibility of the fixations, we visualize them in Fig.\ref{fig:vis_attn}(a), where they are drawn from the attention maps generated by the models.
As illustrated in Fig.\ref{fig:attn}, fixations are sampled from the models' attention maps, and the inhibition of return (IOR) fosters models to explore new areas of images. The first four fixations are marked with red, blue, green, and black squares in order. 
Fig.~\ref{fig:vis_attn}(a) demonstrates that the learned fixation points are placed on the foreground objects as opposed to focusing on background regions with no information.

\begin{table}[]
  \caption{Attack success rates of the models.}
  \label{table:adv}
  \centering
\begin{tabular}{@{}ccccccccccc@{}}
\toprule
\multirow{2}{*}{Model} &      & \multicolumn{4}{c}{Untargeted PGD}                                        &  & \multicolumn{4}{c}{Targeted PGD}                                         \\ \cmidrule(lr){3-6} \cmidrule(l){8-11} 
                       & $\epsilon$: & 2e-3            & 3e-3            & 5e-3            & 7e-3            &  & 3e-3           & 5e-3            & 7e-3            & 1e-2            \\ \midrule
Retina                 &      & \textbf{28.2\%} & \textbf{42.4\%} & \textbf{65.2\%} & \textbf{78.8\%} &  & \textbf{4.7\%} & \textbf{13.6\%} & \textbf{31.7\%} & \textbf{55.2\%} \\
Crop-D                 &      & 49.2\%          & 73.2\%          & 89.1\%          & 96.1\%          &  & 22.4\%         & 56.6\%          & 85.1\%          & 96.1\%          \\
Crop-S                 &      & 65.1\%          & 81.1\%          & 93.0\%          & 97.8\%          &  & 31.2\%         & 70.7\%          & 88.6\%          & 95.9\%          \\
S3TA                   &      & 94.4\%          & 97.1\%          & 99.8\%          & 99.9\%          &  & 72.2\%         & 93.2\%          & 99.2\%          & 100.0\%         \\
FF-CNN                 &      & 91.5\%          & 96.5\%          & 99.8\%          & 99.8\%          &  & 82.8\%         & 98.8\%          & 99.8\%          & 99.8\%          \\ \bottomrule
\end{tabular}
\end{table}

\begin{table}[]
  \caption{Attack success rates from FGSM and SPSA.}
  \label{table:adv1}
  \centering
\begin{tabular}{@{}ccccccccccc@{}}
\toprule
\multirow{2}{*}{Model} &      & \multicolumn{4}{c}{FGSM}                                        &  & \multicolumn{4}{c}{SPSA}                                         \\ \cmidrule(lr){3-6} \cmidrule(l){8-11} 
                       & $\epsilon$: & 2e-3            & 3e-3            & 5e-3            & 7e-3            &  & 7e-3           & 1e-2            & 2e-2            & 3e-2            \\ \midrule
Retina                 &      & \textbf{21.7\%} & \textbf{31.7\%} & \textbf{42.9\%} & \textbf{52.6\%} &  & \textbf{5.7\%} & \textbf{13.5\%} & \textbf{21.5\%} & \textbf{27.3\%} \\
Crop-D                 &      & 33.3\%          & 44.1\%          & 59.8\%          & 65.8\%          &  & 21.5\%         & 26.7\%          & 46.5\%          & 61.9\%          \\
Crop-S                 &      & 52.2\%          & 64.6\%          & 77.9\%          & 86.5\%          &  & 65.9\%         & 77.5\%          & 89.5\%          & 96.8\%          \\
S3TA                   &      & 67.9\%          & 76.5\%          & 87.0\%          & 90.6\%          &  & 100.0\%         & 100.0\%          & 100.0\%          & 100.0\%         \\
FF-CNN                 &      & 82.0\%          & 88.2\%          & 93.5\%          & 92.0\%          &  & 100.0\%         & 100.0\%          & 100.0\%          & 100.0\%          \\ \bottomrule
\end{tabular}
\end{table}

\subsection{Adversarial Attacks}
Evaluating the robustness of computer vision models against adversarial attacks is crucial for understanding their resilience and reliability in real-world applications. Adversarial attacks aim to manipulate model predictions by introducing carefully crafted perturbations to the input images. These attacks can be broadly classified into two types: targeted and untargeted. In a targeted attack, the goal is to deceive an image-computable model, which correctly classifies an image, into labeling it as a different, attacker-specified class. In contrast, an untargeted attack strives to reduce the likelihood of the model recognizing the image as its original class, causing it to classify the image incorrectly.

We evaluate the adversarial robustness of the models trained on ImageNet100 using attack success rate (ASR), which represents the ratio of successful attacks to total attack attempts. A model with a lower ASR is deemed more resistant to attacks. 
For attack algorithms, projected gradient descent (PGD) with \(L_{\infty}\) is used while varying the maximum perturbation budget allowed ($\epsilon$). We iterate PGD for 100 times with the step size at \(\epsilon/20\). 
We also consider Fast Gradient Signed Method (FGSM) attack (\cite{goodfellow2014explaining}) and SPSA attack (gradient-free) (\cite{uesato2018adversarial}). 
All the sampling methods (Retinal transformation and image crop) are fully differentiable.

To understand how these attack strategies impact the dynamical nature of our models, we further explore their responses under both targeted and untargeted attack scenarios.
The dynamical models (Crop-S, Crop-D, Retina, and S3TA) are set to take twelve gazes at each image under the untargeted/targeted attack to minimize/maximize the prediction probability of the true/target class from all gazes. 
To reflect the stochastic nature of eye movements (\cite{burak2010bayesian, kuang2012temporal}), Gaussian noise (with $0$ mean and $0.1$ standard deviation) is added to the fixation points generated from the models (Crop-S, Crop-D and Retina), and Expectation over Transformation (EOT) (\cite{athalye2018synthesizing}) is used to deal with the randomness. In our experiments, we average the gradients over $40$ iterations. Considering the huge amount of computational costs from $100$ iterations of PGD and $40$ iterations from EOT, we further sample the validation set of ImageNet100 to include $1,000$ images to reduce computational costs. For an SPSA attack, we use a sample size of $4096$.

\begin{figure}
  \centering
  \includegraphics[width=1.0\linewidth]{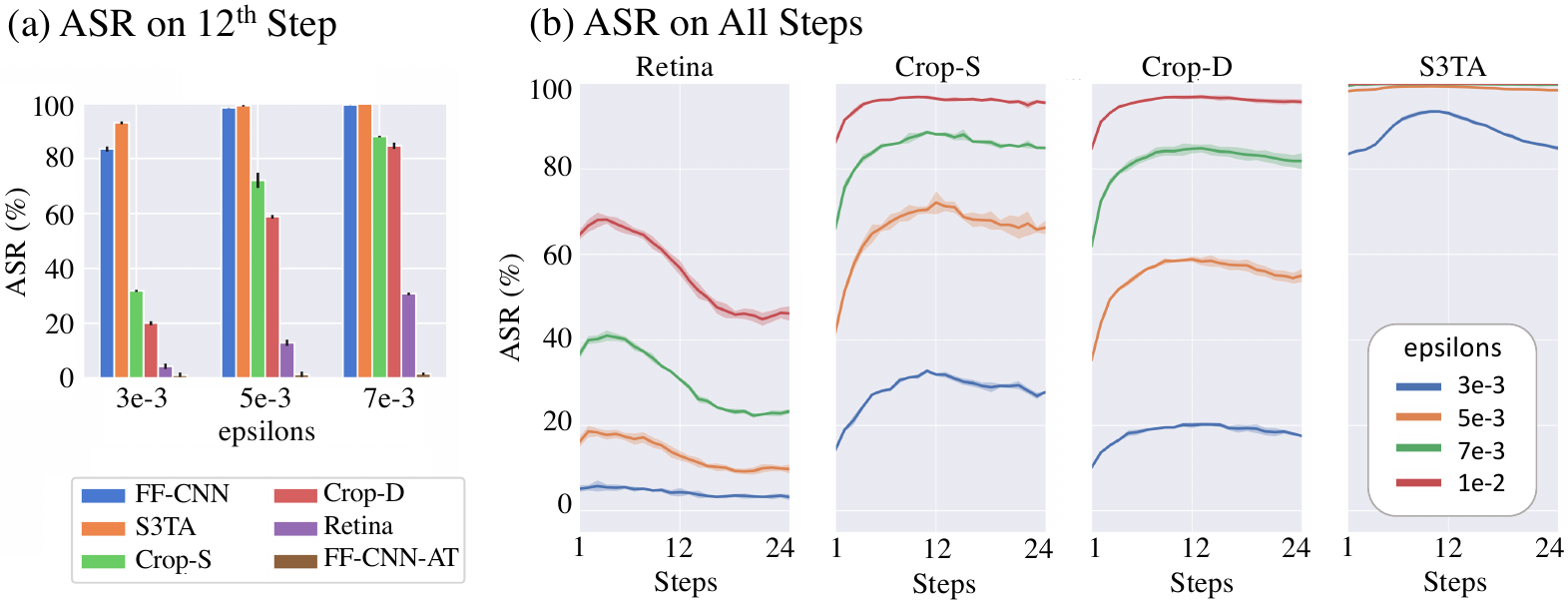}
  \caption{
  Results of targeted PGD attack. (a) Attack success rate (ASR) from $12$th step. (b) ASR from all steps. 
  }
  \label{fig:adv_results}
\end{figure}

\subsubsection{Effects of the retinal transformation}
Table~\ref{table:adv} shows the ASR of dynamical models at the \(12\)-th step given adversarial examples with various adversarial noise levels, epsilon (\(\epsilon\)). The result shows that the models with overt attention (Crop-S, Crop-D and Retina) are more robust than the covert attention (S3TA). Table~\ref{table:adv1} shows the results from FGSM attack and SPSA attack. 
Consistent with previous findings, we observe that combining multi-resolution patches (Crop-D) is more robust than a single-resolution patch (in Crop-S)  (\cite{vuyyuru2020biologically}).
The model Retina maintains the highest robustness compared to other baseline models, and outperforms Crop-S and Crop-D by a large margin. This result suggests that non-uniform retinal transformation is a key mechanism for adversarial robustness among the three sampling strategies.  
In general, models using explicit fixations (Crop-S, Crop-D, and Retina) demonstrate greater robustness compared to S3TA, whose robustness is relatively low and akin to that of FF-CNN when adversarial training is not applied.

It should be noted that the performance of the S3TA model in our study, particularly its attack success rate, differs from what was reported in the original paper. This discrepancy arises from our focus on architectural design choices rather than learning strategies, such as adversarial training. In our experiments, we employed S3TA to contrast the effects of overt attention (eye movements) and covert attention (soft attention), without the adversarial training used in the original study. As a result, the S3TA model in our research, lacking this adversarial training component, exhibits a higher attack success rate than its counterpart in the original study.

\subsubsection{Effects of the number of recurrent steps}
Unlike feed-forward CNNs, dynamical models can vary and unroll their computational graphs with increasing time. Hence, we evaluate these models' ASR as a function of time at each recurrent inference step. 
Fig.~\ref{fig:adv_results}(a) visualizes ASR at the $12$-th step of the targeted attack from Table~\ref{table:adv} including adversarially trained FF-CNN (FF-CNN-AT), which achieves highest robustness. 
 Fig.~\ref{fig:adv_results}(b) shows ASR for all $24$ steps from the targeted PGD attack. All models except for the model Retina appear to maintain or increase ASR by taking more glances for object recognition. In contrast, the model Retina increases its robustness during the steps both under ($1\sim 12$ steps) and after ($12\sim 24$ steps) the attack.

\begin{figure}
  \centering
  \includegraphics[width=.99\linewidth]{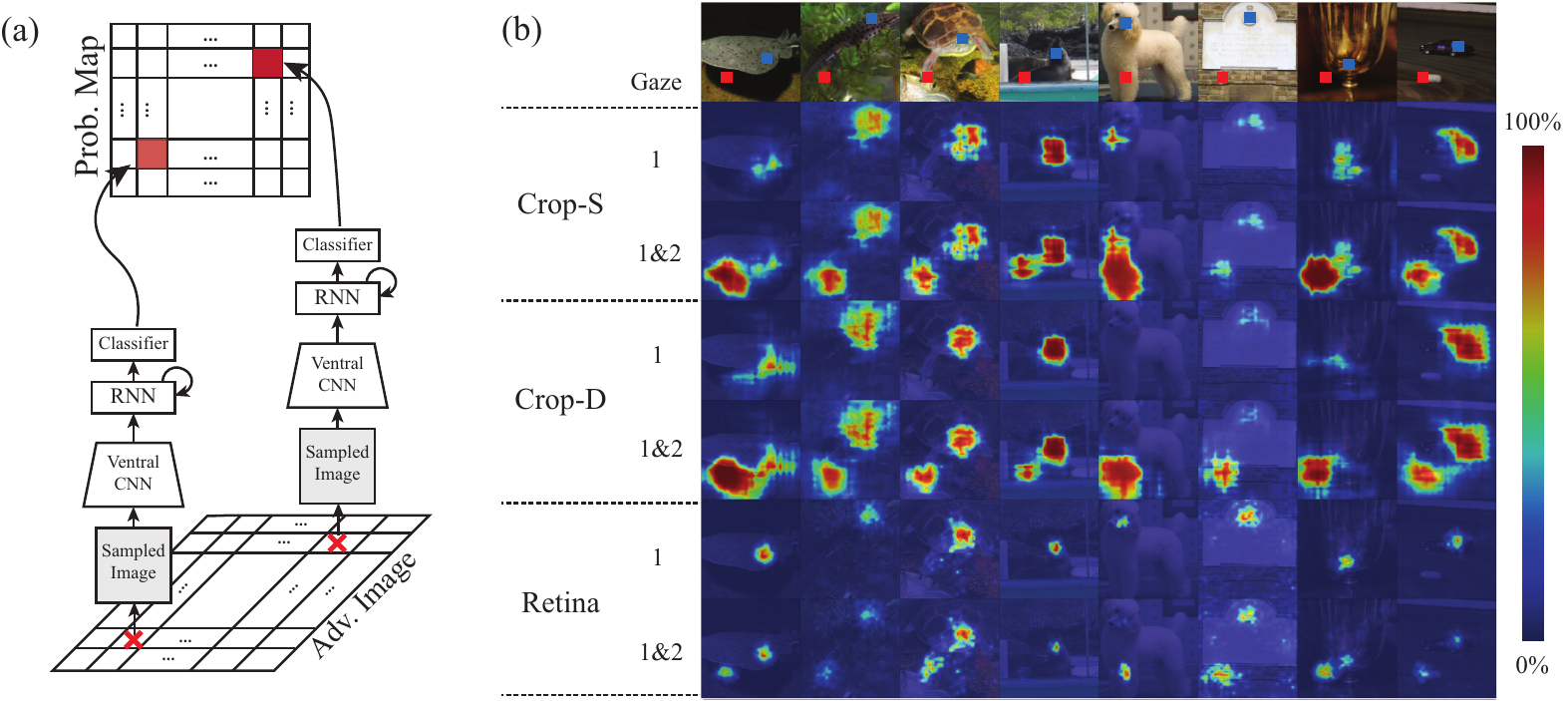}
  \caption{
  (a) Methodology for Generating Adversarial Probability Maps: These maps are created by systematically altering the fixation point across all regions of the image and recording the model's output probability of identifying the target class used in a targeted adversarial attack scenario. 
  (b) Visualization of the Experiment: The first row shows original images with fixations used during attack generation (blue dot: first fixation, red dot: second fixation). Subsequent rows present adversarial probability maps for each model under two scenarios during the attack: 1) using only the first fixation, and 2) incorporating both first and second fixations. These scenarios reflect different levels of exposure to adversarial influence. The color gradient on the maps signifies the model’s likelihood of misclassifying the image as the adversarial target class based on the fixation: warmer colors (towards red) suggest higher probabilities, and cooler colors (towards blue) for lower probabilities. This indicates how the fixation strategy impacts the model’s vulnerability to adversarial attacks.
  }
  \label{fig:advmap}
\end{figure}

\subsubsection{Effect of sequential eye movements}
To intuitively understand this difference, we visualize the regions that are affected by the adversarial attack for each model (Fig.~\ref{fig:advmap}). 
For each model (Crop-S, Crop-D, and Retina), we generated two sets of adversarial images: the first set using only the model's initial, learned fixation point, reflecting the model's natural point of focus in an adversarial context; and the second set incorporating an additional, artificially designated fixation, allowing us to examine the impact of a two-step fixation sequence. In this sequence, the first adversarial image is obtained using the model's natural first fixation, while the second image employs both the original and a new, predefined fixation in the lower left corner, effectively simulating a more complex viewing pattern.
The second fixation is manually selected at the lower left corner to ensure that the first fixation point to be located in the periphery of the second fixation. 
The first (blue) and the second (red) fixations are marked on image examples in the first row of  Fig.~\ref{fig:advmap}(b). Two adversarial images from each model are generated under the targeted PGD attack: one deploying only the first fixation (blue), and the other deploying both fixations. The resulting adversarial images are not shown in Fig.~\ref{fig:advmap}. 
The same Gaussian noise and the targeted attack used for Table~\ref{table:adv} and Fig.~\ref{fig:adv_results} are used again for the experiment.

Once the adversarial images are generated, the models take them as input and perform a recognition task by making fixations on all the possible locations on the images as illustrated in Fig.~\ref{fig:advmap}(a). 
In this experiment, our goal is to assess the model's response to adversarial attacks by analyzing its performance across a range of potential fixation points. To this end, we overlay a $32 \times 32$ virtual grid over each adversarial image, as depicted at the bottom of Fig.~\ref{fig:advmap}(a). This grid represents an array of possible fixation locations, allowing us to simulate how the model might focus on different parts of the image. Each grid point serves as a hypothetical fixation site, enabling a methodical exploration of the model’s responses to a diverse set of focal points.

For each grid location, we direct the model's fixation to that point and record its prediction probability for a randomly selected target class used in the targeted attack. These probabilities are mapped to the corresponding locations on the grid, as shown in Fig.~\ref{fig:advmap}(a), top. This approach provides us with a detailed perspective on how the model processes adversarial images, revealing its susceptibility or resilience to attacks based on where it focuses.

Each fixation is treated as an isolated event in this experiment, meaning the recurrent neural network (RNN) in the model's ventral stream is reset for every new fixation. This enables us to consider the model's object recognition capabilities as a series of independent, one-step tasks. By systematically varying the fixation points across the grid and observing the model's classification outcomes, we gain valuable insights into how its visual attention and processing strategies influence its robustness against adversarial attacks.

Fig.~\ref{fig:advmap}(b) displays the probability maps for the target class from the models. In these maps, color indicates the likelihood of the models classifying the image as the targeted class in the adversarial attack, with red signifying a higher probability and blue a lower one. This means that fixations placed on red regions in an adversarial image are likely to lead the model to classify the image as the target class. Conversely, fixations on blue regions tend to result in the model not classifying the image as the target class. Thus, the probability maps effectively represent the model’s vulnerability to the adversarial attack, considering all possible fixation locations on the image.

As shown in Fig.~\ref{fig:advmap}(b), the locations near the attacked fixation points are more vulnerable. However, although the fixations are the same across all models, the area that is highly affected by the adversarial attack is different for distinct models. The Retina model shows the smallest area affected by the adversarial attack compared to the Crop-S and Crop-D models. By comparing the probability maps from a single fixation and two fixations, we observe that the attacked area near the first fixation is persistently affected during the second attack. However, for the Retina model, the area affected by the attack at the first fixation is reduced after including the second fixation.

This observation can be attributed to the fact that the convolutional operation on the retinal images does not maintain the property of translational equivariance, while the convolution on the images in a regular grid does. When the fixation in Crop-S or Crop-D changes a little bit, the representations after convolutional operations will be simply shifted to the new fixation location. However, for the Retina model, small changes in the fixation will result in large changes in the retinal images after foveation and non-uniform sampling. This may further alter its representation significantly in the stacked CNN layers. This fixation-dependent representation is not a simple translation but a change of the whole layout. Therefore, the adversarial noise generated at the certain fixation does not necessarily fool the Retina model when a fixation point is made on other locations. 

This is in line with the human brain, where foveal vision and peripheral vision have different functional roles. The foveal vision focuses on object recognition, while the peripheral vision has better adversarial robustness (\cite{logothetis1995shape, riesenhuber1999hierarchical, harrington2021finding}). In Fig.~\ref{fig:advmap}, when the first fixation is made on the object, the adversarial noise is calculated based on the object in the foveal vision. Then when the second fixation is set at the corner, the object is placed in the peripheral vision. Because of this difference, the adversarial noise from the foveal vision (from the first fixation) fools the model less effectively when it is placed on the peripheral vision (from the second fixation).

\section{Discussions}
This study introduces brain-inspired mechanisms into computer vision models, specifically focusing on retinal sampling, adaptive eye movement, and recurrent processing. Our models can produce more human-like visual attention and be more robust against adversarial attacks. Through computational experiments, we demonstrate that 1) visual attention emerges as a strategy that optimizes eye movement for efficient object recognition, 2) non-uniform retinal sampling and varying fixations contribute to adversarial robustness.  

We found that the Retina model and the Crop-D model exhibit superior performance in mimicking human visual attention compared to baseline models. A key feature of these models is their fovea/periphery approach, which involves high-resolution foveal viewing combined with low-resolution peripheral processing. This approach mirrors the human vision system where both foveal and peripheral views are integral for attention. The high-resolution foveal view in these models allows for detailed analysis of specific regions of interest, paralleling the human use of fovea to focus on and scrutinize central vision details. In contrast, the low-resolution peripheral processing effectively manages the broader scene context, assisting in overall interpretation and situational awareness, similar to human peripheral vision.

This dual approach to visual processing, inspired by human vision, enables these models to effectively learn and prioritize visual attention. By dynamically shifting focus between detailed and broader views, the models more accurately identify relevant image areas, enhancing their ability to mimic human attention patterns. This capability aligns with the concept of attention mining (\cite{wei2017object, li2018tell}) in computer vision, wherein the models, akin to human visual processing, distribute attention across a wider range of relevant features, reducing biases towards certain dominant features. The retina model, in particular, excels in this aspect by adaptively focusing on various parts of an image, thus offering a more balanced and comprehensive analysis.

The implications of these findings are significant, suggesting that integrating elements of human visual processing, especially the balance between foveal and peripheral attention, can create more intuitive and effective AI vision systems. These systems would excel not only in recognizing fine details but also in understanding the broader context of a scene. Such advancements hold promise for applications in areas like autonomous navigation, surveillance, and human-computer interaction, where nuanced visual perception is crucial.

In this study, models incorporating retinal transformation have emerged as the most robust among all tested. This superior robustness can be attributed to retinal sampling, which complicates adversarial attacks. Adversarial perturbations tailored for a specific retinal pattern, associated with a given fixation, prove difficult to generalize across different retinal patterns. This is because as the eyes move and fixate on different locations, the retinal patterns of the same objects are represented with varying features, resolutions, and scales. 
Unlike models that process images on a regular grid, where perturbations at one location can easily be transferred to other areas, retinal sampling introduces an element of unpredictability for attackers, thereby enhancing robustness.

Building on the advantages of retinal transformation, models with sequential eye movement further augment this robustness. In scenarios without time constraints, such models have the capacity to scan an image multiple times, each time attending to a different region. This adaptability, especially when eye movements are stochastic, introduces an additional layer of complexity for attackers. If attackers lack knowledge of the model's subsequent fixation points, our experiments suggest that the model can utilize additional glances to rectify errors induced by adversarial images. Notably, models with overt eye movement, or overt attention, consistently demonstrate greater robustness compared to the model with covert attention (S3TA). This finding underscores the significance of variable attention mechanisms in enhancing model resilience. The integration of these brain-inspired approaches – retinal transformation and sequential eye movements – not only bolsters the models' defenses against adversarial attacks but also aligns their functioning more closely with human visual processing. Such advancements hold promise for future developments in AI, potentially leading to more intuitive, efficient, and resilient computer vision systems.

Building on the strengths and insights gained from our research, it is important to also acknowledge some of the limitations and areas for further exploration. In advancing the work of Vuyyuru et al. (2020), we integrated an attention mechanism in our models that learns and adapts fixation points dynamically. This approach marks a departure from the static fixation strategies used in prior studies. However, our investigation into the impact of this learned fixation strategy on model robustness revealed complex dynamics. We observed no significant difference in robustness between learned and random or predefined fixations (results are not included in the main text). However, a noticeable decrease in robustness occurred when models operated without the Inhibition of Return (IOR), suggesting the importance of varied and shifting fixation points in enhancing adversarial robustness.

The influence of the dataset characteristics on our findings is also noteworthy. Our evaluations primarily utilized the ImageNet dataset, characterized by images with large, prominent objects. In such a context, both random and learned fixations often focus on these dominant objects, reducing the distinction between the two fixation strategies. A clearer difference might emerge with datasets featuring smaller objects, where the precision of fixations could play a more significant role. Absence of IOR led to static fixations, allowing for the persistence of adversarial patterns. In contrast, models with IOR or adaptive fixation strategies, which involve dynamic eye movements, showed an ability to mitigate the impact of adversarial patterns over time, enhancing overall robustness.

In our exploration of brain-inspired mechanisms within computer vision models, we have only scratched the surface of potential applications. While our study primarily focused on retinal transformation and adaptive eye movements, other mechanisms prevalent in the brain, such as top-down and recurrent connections, hold significant promise. In our models, recurrent connections were utilized at the end of the \texttt{VentralCNN}, but top-down connections, crucial for integrating higher-level cognitive processes with sensory inputs, were not implemented. Delving deeper into these mechanisms could pave the way for developing more sophisticated and resilient computer vision models. This exploration has the potential not only to enhance AI technology but also to contribute to our understanding of human vision processes and behaviors, bridging the gap between computational modeling and biological reality.

It is also essential to address some limitations present in our study. One notable area is the object recognition accuracy of models using retinal transformation, particularly in single-label classification tasks. This approach, while mimicking the human eye's foveal-peripheral dynamics, introduces a non-linear distortion to images. The high-resolution focus in the fovea and significant downsampling in peripheral areas can affect the efficiency of convolutional operations, especially since the same kernels process regions of varying scale. Consequently, our model exhibits suboptimal performance compared to other baseline methods when dealing with large objects that extend into the peripheral field. Conversely, for images with smaller objects or more complex scenes, the retinal transformation demonstrates superior performance, underscoring its suitability for handling intricate visual information.

\section{Acknowledgements}
This research is supported by the Collaborative Research in Computational Neuroscience (CRCNS) program from National Science Foundation (Award\#: IIS 2112773) and the University of Michigan.

\subsection*{Appendix A. Foveated Imaging}

In the foveated image, the foveated region maintains high acuity, but the periphery has reduced acuity. To implement the eccentricity-dependent acuity, we apply different levels of Gaussian blur based on the eccentricity. Here, we assume that the fixation is at the center of an image. Similar procedures can be generalized to non-centered fixation. 

To illustrate a varying blurring effect, we use three 2D isotropic Gaussian kernels, \(\bm{K}_{b1}\), \(\bm{K}_{b2}\) and \(\bm{K}_{b3}\). The kernel \(\bm{K}_{bi}\) ($i \in [1, 2, 3]$) is in size $7\times 7$ and have a variance \(\sigma_{bi}^2\). The \(\sigma_{b1}\), \(\sigma_{b2}\) and \(\sigma_{b3}\) are set to 1, 3 and 5, respectively in this study. 

\begin{figure}
  \centering
  \includegraphics[width=0.9\linewidth]{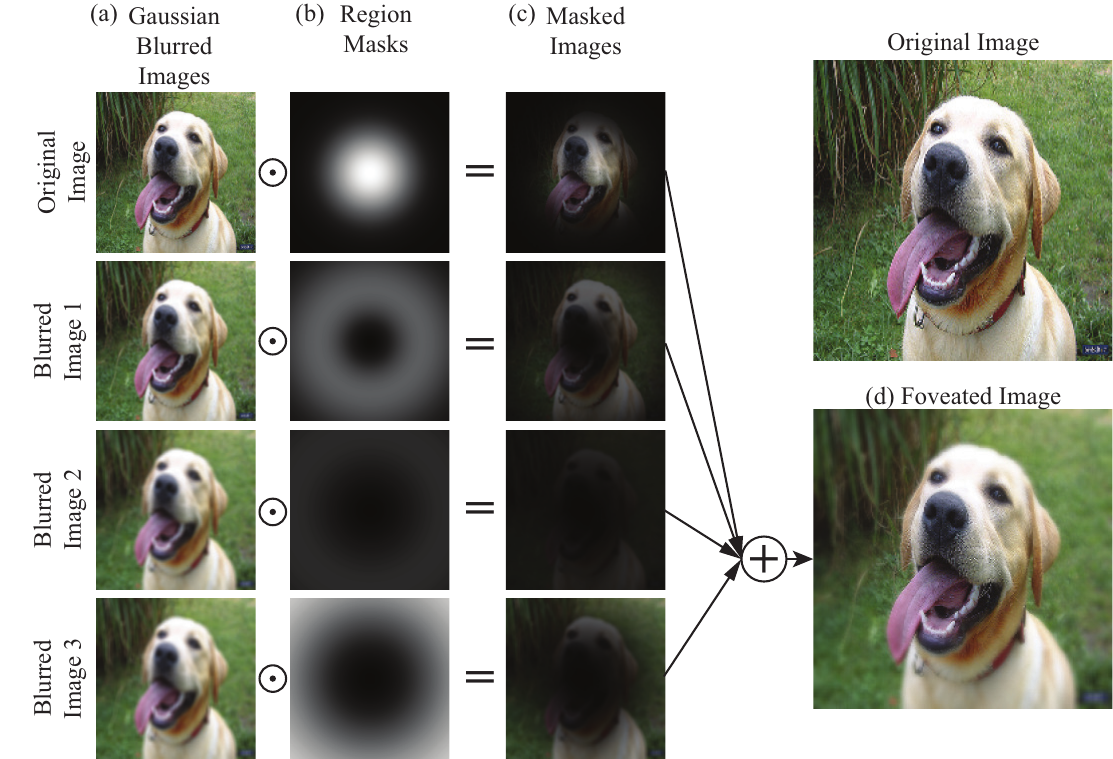}
  \caption{Illustration of the foveated image generated from an example image, when the fixation is at the image center. (a) The original image is blurred by applying different Gaussian kernels, (b) region masks applied to the blurred images. The black and white colors represent the range of values ($0$, $1$). (c) Results of element-wise multiplication of (a) and (b). (d) The foveated image by summing up the masked images.
}
  \label{figs:fovimg}
\end{figure}

Firstly, the Gaussian kernels \(\bm{K}_{b1}\), \(\bm{K}_{b2}\) and \(\bm{K}_{b3}\) are separately applied to the original image, resulting in blurred images, \(\bm{A}_{b1}\), \(\bm{A}_{b2}\) and \(\bm{A}_{b3}\). The original image (\(\bm{A}\)) and the blurred images are shown in Fig. \ref{figs:fovimg}(a). 

Secondly, region masks are produced. The region masks are used to confine the regions with a specific level of blurring. The masks appear as concentric rings. They act as filters to pass the image with the desired resolution to the target area. Applying such masks to the blurred images makes the blurring dependent on the eccentricity. For a foveated image, a region mask designed to pass the fovea region is applied to an image with original resolution (Fig. \ref{figs:fovimg}(b) top row). For the periphery, region masks for the peripheral regions are applied to the blurred images (shown in Fig. \ref{figs:fovimg}(b) $2$nd, $3$rd and $4$th row from the top).

To generate region masks, 2D isotropic Gaussian kernels are first generated. The Gaussian kernels are denoted as \(\bm{T}_{t1}\), \(\bm{T}_{t2}\) and \(\bm{T}_{t3}\), and they are in size \(224\times 224\). \(\bm{T}_{t1}\), \(\bm{T}_{t2}\) and \(\bm{T}_{t3}\) have means at the fixated location and variances \(\sigma_{t1}^2\), \(\sigma_{t2}^2\) and \(\sigma_{t3}^2\), respectively (In this study, \(\sigma_{t1}=40\), \(\sigma_{t2}=70\), \(\sigma_{t3}=90\)). The generated Gaussian kernels are normalized to make their maximal amplitude $1.0$. The region masks \(\bm{M}_{1}\), \(\bm{M}_{2}\), \(\bm{M}_{3}\) and \(\bm{M}_{4}\) are generated as Eq. \ref{eqn:r1}-\ref{eqn:r4}

\begin{subequations}
\begin{align}
    \bm{M}_{1} &= \bm{T}_{t1}  \label{eqn:r1}\\
    \bm{M}_{2} &= \bm{T}_{t2} - \bm{T}_{t1}  \label{eqn:r2}\\
    \bm{M}_{3} &= \bm{T}_{t3} - \bm{T}_{t2} \label{eqn:r3}\\
    \bm{M}_{4} &= \bm{1} - \bm{T}_{t3}  \label{eqn:r4}
\end{align}
\label{eqn:regionmasks}
\end{subequations}

where \(\bm{1}\) is a 2D matrix filled with ones whose size is identical to \(\bm{T}_{t3}\)
The region masks are shown in Fig. \ref{figs:fovimg}(b) and it can be checked that each region mask passes distinct image regions based on the eccentricity. 

Once the blurred images and the region masks are obtained, the region masks are applied to the corresponding original and blurred images as element-wise multiplications (Eq. \ref{eqn:m1}-\ref{eqn:m4}). 
\begin{subequations}
\begin{align}
    \bm{A}_{m1} &= \bm{A} \odot \bm{T}_{t1}    \label{eqn:m1}\\
    \bm{A}_{m2} &= \bm{A}_{b1} \odot \bm{T}_{t2} \label{eqn:m2}\\
    \bm{A}_{m3} &= \bm{A}_{b2} \odot \bm{T}_{t3} \label{eqn:m3}\\
    \bm{A}_{m4} &= \bm{A}_{b3} \odot \bm{T}_{t4} \label{eqn:m4}
\end{align}
\label{eqn:maskedimg}
\end{subequations}

where the operation \(\odot\) is an element-wise multiplication. The first region mask \(\bm{T}_{t1}\) designed to pass the center of fixation is applied to the original image so that the foveated image maintains high acuity in the fixated region. On the contrary, the region masks assigned to the peripheral regions are applied to the corresponding blurred images. The masked images are shown in Fig. \ref{figs:fovimg}(c).

Lastly, the masked images are summed together to produce the foveated image. 

\begin{equation}
    \text{Foveated Image} = \bm{A}_{m1} + \bm{A}_{m2} + \bm{A}_{m3} + \bm{A}_{m4}
\end{equation}

The resulting image is shown in Fig. \ref{figs:fovimg}(d). In the foveated image, the region near the fixation maintains the high acuity, and the periphery regions are gradually blurred as the eccentricity grows.

\subsection*{Appendix B. ImageNet100}
The randomly selected $100$ classes from $1000$ classes of ILSVRC2012 (\cite{deng2009imagenet}) are: 

\noindent [n01496331  n01756291  n01833805  n02025239  n02100583  n02137549  n02480495  n02808440  n03124043  n03291819  n03770679  n03902125  n04201297  n04371430  n09246464
n01531178  n01768244  n01843383  n02028035  n02102480  n02138441  n02492660  n02834397  n03127747  n03388183  n03781244  n03956157  n04251144  n04399382  n09472597
n01630670  n01797886  n01847000  n02077923  n02105412  n02172182  n02504458  n02892201  n03131574  n03443371  n03785016  n04037443  n04275548  n04505470
n01644900  n01806143  n01871265  n02087046  n02106166  n02190166  n02640242  n02963159  n03180011  n03494278  n03796401  n04040759  n04325704  n04536866
n01667778  n01807496  n01872401  n02088632  n02108089  n02233338  n02701002  n02971356  n03216828  n03662601  n03837869  n04049303  n04335435  n04589890
n01669191  n01817953  n01968897  n02089078  n02113712  n02317335  n02787622  n02977058  n03249569  n03710193  n03877472  n04146614  n04355338  n04591713
n01694178  n01824575  n01980166  n02094258  n02119789  n02410509  n02795169  n03100240  n03272010  n03733131  n03877845   n04162706  n04356056  n07614500]


%

\end{document}